\documentclass[letterpaper]{article} % DO NOT CHANGE THIS
\usepackage[draft]{aaai2026}  % DO NOT CHANGE THIS
\usepackage{times}  % DO NOT CHANGE THIS
\usepackage{helvet}  % DO NOT CHANGE THIS
\usepackage{courier}  % DO NOT CHANGE THIS
\usepackage[hyphens]{url}  % DO NOT CHANGE THIS
\usepackage{graphicx} % DO NOT CHANGE THIS
\usepackage{natbib}  % DO NOT CHANGE THIS AND DO NOT ADD ANY OPTIONS TO IT
\usepackage{caption} % DO NOT CHANGE THIS AND DO NOT ADD ANY OPTIONS TO IT
\usepackage{algorithm}
\usepackage{algorithmic}
\usepackage{caption} % DO NOT CHANGE THIS AND DO NOT ADD ANY OPTIONS TO IT
\usepackage{amsmath} % Added to support align environment
\usepackage{newfloat}
\usepackage{listings}
\DeclareCaptionStyle{ruled}{labelfont=normalfont,labelsep=colon,strut=off} % DO NOT CHANGE THIS
\floatstyle{ruled}
\newfloat{listing}{tb}{lst}{}
\floatname{listing}{Listing}
\usepackage{booktabs}
\usepackage{multirow}
\usepackage{xcolor}  % 加在导言区
\usepackage{tabularx}
\usepackage{cleveref}
\usepackage{xspace}
\usepackage{amssymb} % Required for \mathbb

\makeatletter
\DeclareRobustCommand\onedot{\futurelet\@let@token\@onedot}
\def\@onedot{\ifx\@let@token.\else.\null\fi\xspace}

\def\eg{\emph{e.g}\onedot} 
\def\ie{\emph{i.e}\onedot}

\makeatother

\title{An Empirical Study on How Video-LLMs Answer Video Questions}
\author{
     Chenhui Gou\textsuperscript{\rm 1}, Ziyu Ma\textsuperscript{\rm 2, 1}, Zicheng Duan\textsuperscript{\rm 3}, Haoyu He\textsuperscript{\rm 1}, Feng Chen\textsuperscript{\rm 3}, Akide Liu\textsuperscript{\rm 1}, \\
     Bohan Zhuang\textsuperscript{\rm 4}, Jianfei Cai\textsuperscript{\rm 1}, Hamid Rezatofighi\textsuperscript{\rm 1}
}
\affiliations{
    \textsuperscript{\rm 1}Monash University \textsuperscript{\rm 2}Hunan University \textsuperscript{\rm 3}The University of Adelaide \textsuperscript{\rm 4}Zhejiang University}

\begin{document}

\maketitle

\begin{abstract}
Taking advantage of large-scale data and pretrained language models, Video Large Language Models (Video-LLMs) have shown strong capabilities in answering video questions. However, most existing efforts focus on improving performance, with limited attention to understanding their internal mechanisms. This paper aims to bridge this gap through a systematic empirical study. To interpret existing VideoLLMs, we adopt attention knockouts as our primary analytical tool and design three variants: Video Temporal Knockout, Video Spatial Knockout, and Language-to-Video Knockout. Then, we apply these three knockouts on different numbers of layers (window of layers). By carefully controlling the window of layers and types of knockouts, we provide two settings: a global setting and a fine-grained setting. Our study reveals three key findings: (1) Global setting indicates Video information extraction primarily occurs in early layers, forming a clear two-stage process—lower layers focus on perceptual encoding, while higher layers handle abstract reasoning; (2) In the fine-grained setting, certain intermediate layers exert an outsized impact on video question answering, acting as critical outliers, whereas most other layers contribute minimally; (3) In both settings, we observe that spatial-temporal modeling relies more on language-guided retrieval than on intra- and inter-frame self-attention among video tokens, despite the latter’s high computational cost. Finally, we demonstrate that these insights can be leveraged to reduce attention computation in Video-LLMs. To our knowledge, this is the first work to systematically uncover how Video-LLMs internally process and understand video content, offering interpretability and efficiency perspectives for future research.
\end{abstract}

%%%% intro

\section{Introduction}\label{sec:intro}
\begin{figure*}[t]
    \centering
    \includegraphics[width=1\linewidth]{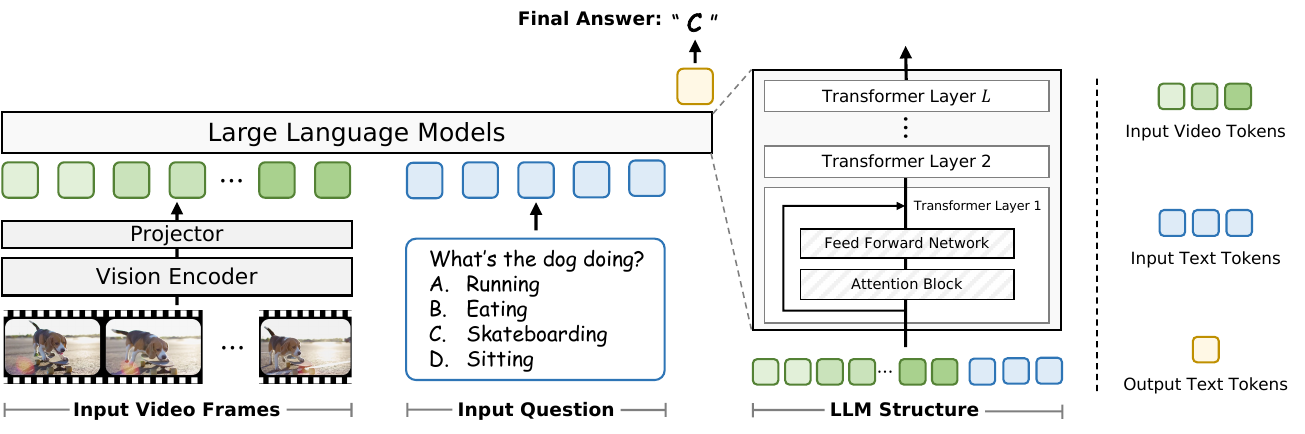}
    \caption{Typical architecture design for Video-LLMs.}
    \label{fig:1}
\end{figure*}
\noindent Video Large Language Models (Video-LLMs) have recently shows their strong ability to understand video content and answer various types of questions \cite{zhang2024llavanextvideo,zhang2024videoinstructiontuningsynthetic,yao2024minicpm,Qwen2.5-VL,chen2024expanding,cheng2024videollama}.
Most recent research on Video-LLMs focuses on improving model performance, e.g.,
enhancing the quality and scale of video instruction datasets \cite{zhang2024videoinstructiontuningsynthetic,yao2024minicpm,Qwen2.5-VL,chen2024expanding,cheng2024videollama}, extending the input frame length \cite{zhang2024long,xue2024longvila,liu2024kangaroo}, and refining positional encoding for video tokens \cite{liu2025vrope,ge2024v2pe,wei2025videorope}. However, there remains limited understanding of their internal mechanisms—specifically, how they process and reason about video content. Gaining deeper insights into these mechanisms is crucial for enhancing explainability, improving model efficiency, and facilitating future model developments.

In the image domain, many researchers have attempted to increase the interpretability of these large multimodal models to avoid purely black-box usage. These works study how the internal states of MLLMs correspond to their external behaviors \cite{zhang2025cross,kaduri2024s,basu2024understanding,zhao2024first,zhang2024redundancy}. This includes analyses of the information flow from the image to the formation of different stage patterns \cite{zhang2025cross,chen2024image,lin2025boosting}, undesirable content generation patterns in logit distributions \cite{zhao2024first}, two-stage pattern and safety mechanism fine grainedization \cite{xu2024cross}, grounding and evolution of object-related visual cues \cite{neo2024towards,schwettmann2023multimodal,ma2024drvideo}, information storage in model parameters \cite{basu2024understanding}, and reduction of redundant visual tokens \cite{zhang2024redundancy}. Compared to the abundant studies on the interpretability of MLLMs in the image domain, the high-dimensional video domain remains largely unexplored. A recent work \cite{xiao2025videoqa} conducts comprehensive experiments to analyze the behavior of various Video-LLMs and reports several interesting observations: these models excel at VideoQA but falter in temporal grounding, and they are sensitive to language variations while being less sensitive to video perturbations. While this work primarily focuses on analyzing the models' external behaviors, their internal transparency remains largely unexplored.

Our work takes a step toward uncovering the internal patterns of existing Video-LLMs and understanding how these patterns relate to their strong performance in video question answering. Modern Video-LLMs \cite{zhang2024long,wang2025internvideo2,li2024llava,zhang2024llavanextvideo} typically follow a similar architecture: a pre-trained visual encoder that converts video into tokens, a projection layer that maps these tokens into the language space, and a large language model (LLM) that takes both video and language tokens to generate responses. Each video token is processed by every layer of the LLM and interacts with other video tokens and question tokens through attention mechanisms. This facilitates information extraction and propagation across layers, ultimately contributing to the final answer. Within each layer, the attention mechanism can be decomposed into three types: temporal attention (across video frames), spatial attention (within each frame), and language-to-video attention. We design three corresponding attention knockouts that selectively disable one specific type of attention, allowing us to analyze its individual impact. To gain insights at different granularities, we investigate the internal patterns of Video-LLMs under two settings: a \textit{global setting} and a \textit{fine-grained setting} by carefully controlling the knockout layer range and the type of attention knockout. In the global setting, we explore two questions:  
(1) \textit{Do Video-LLMs exhibit a clear two-stage behavior similar to image-level VLMs—i.e., perceptual encoding in early layers and semantic reasoning in later layers—or do they follow a distinct paradigm?}  
(2) \textit{Globally, how does each attention type contribute to video QA performance?}  
In the \textit{fine-grained setting}, we explore:  
(3) \textit{How does each attention type impact VideoQA across different layers?}

We analyze a range of representative Video-LLMs, including LongVA \cite{zhang2024long}, InternVideo2.5 \cite{wang2025internvideo2}, LLaVA-OneVision \cite{li2024llava}, and LLaVA-Video \cite{zhang2024llavanextvideo}, on mainstream VideoQA benchmarks with diverse task types and video lengths. These include short-duration multi-task video QA: MVBench \cite{li2024mvbench}, medium-duration egocentric VideoQA: EgoSchema \cite{mangalam2023egoschema}, and long-duration open-domain VideoQA: Video-MME \cite{fu2024video}. Through extensive experiments comprising over 300 data points, we summarize our key findings:
(1) Observation under global setting: Applying a language-to-video attention knockout starting from a certain proportion of layers (e.g., 60\% of the entire model) does not show a significant impact on performance across various benchmarks and models, as shown in (2) Observation under global setting: Applying a full language-to-video attention knockout leads to a significant performance drop, which is substantially larger than that caused by temporal or spatial knockouts, as shown in \cref{fig:5}.
(3) Observation under fine-grained setting: For most individual layers, language-to-video attention knockouts have a stronger impact than temporal or spatial attention knockouts, as shown in \cref{fig:6}. (4) Observation under fine-grained setting: Knocking out specific layers (e.g., Layers 12–16) results in a substantial performance drop, whereas most other layers have minimal effect, as shown in \cref{fig:6}.

These results reveal the following insights for VideoQA: \textbf{(a)} Video-LLMs exhibit a clear two-stage processing pattern, where early layers primarily focus on extracting video information. \textbf{(b)} Current Video-LLMs heavily rely on language-to-video attention to retrieve and model video content, rather than relying on temporal or spatial video attention—which are more computationally expensive. \textbf{(c)} A small subset of layers plays a critical role in VideoQA and acts as influential outliers in the attention pathway. In the end, we apply the vision token early exit strategy to drop all vision tokens after certain ratio of layers, which directly based on our two-staged finding. We find that this simple strategy can significantly reduce computational overhead with only a minimal impact on performance. To the best of our knowledge, this is the first work to investigate the internal patterns of how Video-LLMs process and understand videos in the context of VideoQA. Our findings enhance the interpretability of Video-LLMs and provide valuable insights for their future development.

\section{Related work}

\begin{figure*}[!htbp]
    \centering
    \includegraphics[width=1\linewidth]{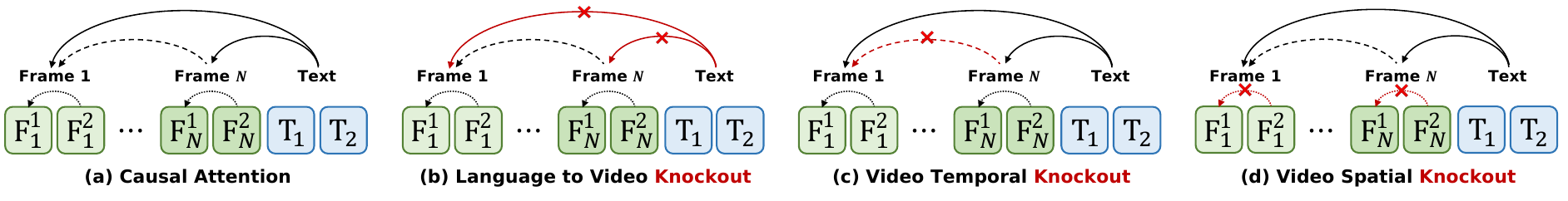}
\caption{
Default causal attention and three types of attention knockout mechanisms. For clarity, we visualize each frame $F_{1 \dots N}$ and all text prompts $T$ using two tokens (\eg, $F_{n}^{\text{1}}, F_{n}^{\text{2}}$) (the actual number of tokens for each frame is larger than 100, and all text prompts contain fewer than 100 tokens). (a) Vanilla Casual Attention: The original causal attention mechanism within LLM models. (b) Language-to-Video Knockout: This mechanism removes attention from text prompts to all tokens in the video frames. (c) Video Temporal Knockout: This approach blocks temporal attention between frames while preserving language-to-video and intra-frame spatial attention. (d) Video Spatial Knockout: This setting disables spatial attention within video frames.}
    \label{fig:2}
\end{figure*}

\noindent\textbf{Video large language models (Video-LLMs).} Existing LLM-based video understanding methods can be categorized into two main types: (1) \textbf{Specialized Video-LLMs.} These methods exploit frozen LLMs (\eg, LLaMA3 \cite{dubey2024llama}, Mistral \cite{jiang2023mistral7b}, and Qwen \cite{yang2024qwen2}) or general-purpose image MLLMs (\eg, BLIP-2 \cite{li2023blip}, LLaVA-Next \cite{liu2024llavanext}, LLaMA-Adapter \cite{zhang2023llama}) while fine-tuning adaptor or LoRA \cite{hu2022lora} modules on specific video understanding datasets. Representative works include FrozenBiLM \cite{yang2022zero}, SeViLA \cite{yu2023self}, LLaMA-VQA \cite{ko2023large}, and Video-LLaMA \cite{zhang2023video}. (2) \textbf{General-purpose Video-LLMs.} These methods are not fine-tuned on specific video understanding datasets but are instead pre-trained on large-scale video datasets, enabling them to handle diverse multimodal tasks such as image/video QA, retrieval, and captioning. They typically enhance the quality and scale of video instruction datasets (\eg, LLaVA-NeXT-Video \cite{zhang2024llavanextvideo}, LLaVA-Video \cite{zhang2024videoinstructiontuningsynthetic}, MiniCPM-V \cite{yao2024minicpm}, Qwen2.5-VL \cite{Qwen2.5-VL}, InternVL2.5 \cite{chen2024expanding}, InternVideo2.5 \cite{wang2025internvideo2}, LLaVA-OneVision \cite{li2024llava}, and VideoLLaMA2 \cite{cheng2024videollama}), extend the input frame length (\eg, LongVA \cite{zhang2024long}, LongVILA \cite{xue2024longvila}, and Kangaroo \cite{liu2024kangaroo}), and refine positional encoding for video tokens (\eg, VRoPE \cite{liu2025vrope}, V2PE \cite{ge2024v2pe}, and VideoRoPE \cite{wei2025videorope}) to improve video understanding capabilities and achieve remarkable performance. In this paper, we examine state-of-the-art and widely used open-source Video-LLMs, including LongVA, InternVideo2.5, LLaVA-OneVision \cite{li2024llava}, and LLaVA-Video \cite{zhang2024videoinstructiontuningsynthetic}, to ensure representative generation ability across different Video-LLMs.

\noindent\textbf{Interpretability of multimodal models.} The interpretability of multimodal models has become a key research focus. Existing approaches can be
roughly classified into three categories: (1) Black-box analysis \cite{cao2020behind,frank2021vision}, which analyzes input-output relationships to understand model behavior, including evaluating the significance of various modalities \cite{cao2020behind} and their contributions to tasks \cite{frank2021vision}; (2) Single-sample attribution \cite{aflalo2022vl,chefer2021generic,lyu2022dime,stan2024lvlm}, which traces predictions back to specific inputs using attention score aggregation \cite{aflalo2022vl,stan2024lvlm}, gradient-based methods \cite{chefer2021generic}, or model disentanglement \cite{lyu2022dime}; and (3) Top-down representation probing \cite{lindstrom2021probing,hendricks2021probing,salin2022vision}, which investigates learned representations to uncover high-level concepts like visual semantics \cite{lindstrom2021probing}, verb understanding \cite{hendricks2021probing}, and shape or size \cite{salin2022vision}. Unlike these approaches, our study investigates the internal processing mechanisms of Video-LLMs in addressing VideoQA tasks.

\noindent\textbf{Mechanistic interpretability of MLLMs.} In image level understanding, several early studies have begun investigating the internal states of MLLMs by connecting external behaviors to specific mechanisms. These aspects encompass information retention within model parameters \cite{basu2024understanding}, unintended content generation reflected in the logit distributions of the initial token \cite{zhao2024first}, the tracking and transformation of object-related visual information \cite{neo2024towards, schwettmann2023multimodal}, the detection of safety mechanisms \cite{xu2024cross}, and the minimization of redundant visual tokens \cite{zhang2024redundancy}. A recent study \cite{xiao2025videoqa} conducts comprehensive experiments to analyze the behavior of various Video-LLMs and shows that these models perform well on VideoQA but lack temporal grounding. Additionally, they are sensitive to language variations while being less affected by video perturbations. While this work primarily focuses on analyzing the models' external behaviors, their internal transparency remains largely unexplored. Our work provides an important initial effort in bridging this gap and serves as a complementary study to \cite{xiao2025videoqa}.

\section{Investigation Design}
\subsection{Preliminary}
A Video-LLM \cite{liu2024llavanext} typically consists of a pre-trained visual encoder, a projection layer, and a decoder-only language model, as shown in Fig. \ref{fig:1}. The visual encoder extracts visual tokens from video inputs, and the projection layer maps them to the language space. Then decoder-only LLM \cite{yang2024qwen2} takes both the video and text tokens and outputs the generated response tokens in an autoregressive manner. Specifically, a video sequence is sampled with $N$ frames, where each frame is encoded into a sequence of visual tokens $\mathbf{F}_i$ using the visual encoder (\eg, CLIP-L-14 \cite{sun2023eva}). These visual tokens are then mapped to the text space via a projection layer.
Mathematically, this process can be expressed as
\begin{equation}
    \mathbf{V} = [\mathbf{F}_i]_{i=1}^{N}, \quad \mathbf{F}_i = \text{Proj}(\text{Enc}_v(\mathbf{x}_i)) \in \mathbb{R}^{d},
\end{equation}
where $\mathbf{x}_i$ represents the $i$-th video frame, $\text{Enc}_v(\cdot)$ denotes the visual encoder, $\text{Proj}(\cdot)$ is the projection layer, and $d$ is the number of hidden dimensions in the LLM. Similarly, the text input is mapped into text embedding tokens $\mathbf{T}$ via a pretrained embedding codebook. The resulting text tokens are concatenated with the video tokens to form the multimodal input sequence:
$\text{MMs}= [\mathbf{F}_1, \dots, \mathbf{F}_N, \mathbf{T}],
$
where $\mathbf{T}$ denotes the text tokens. Please note that this sequence is ordered, starting with video tokens followed by text tokens. We ignore the prefixed system tokens, which are used to control the output behavior of the LLM.
 
\noindent\textbf{Hidden representations and attention.}
The multimodal input sequence is then processed through $L$ layers of transformer blocks \cite{vaswani2017attention} to obtain the hidden representation. Each Transformer block consists of an attention block and a feed-forward network (FFN). The hidden representation at each layer $\ell$ can be written as
\begin{equation}
MMs^{(\ell)} = \mathbf{FFN}(\mathbf{Attn}(MMs^{(\ell-1)})),
\end{equation}
where $\text{Attn}$ represents the attention block. Here we ignore the residual connection for notation simplicity. %The last position text token representation 
The last token of the final layer of the LLM is used to decode the output token.
To understand the video, text tokens need to extract spatial-temporal information from the video. Also, video tokens will communicate with each other within each frame and across frames 
via the attention mechanism.
\\\noindent\textbf{Attention}. 
In each attention block, the multimodal sequence (\( \text{MMs} \)) is projected into the query, key, and value spaces to obtain the \( Q \), \( K \), and \( V \) matrices. Each token in \(\text{MMs}\) exchanges information with all other tokens through causal attention \cite{yang2021causal}.
\begin{align}
\text{CausalAttention}(Q, K, V) &= \text{softmax}\left(\frac{QK^\top}{\sqrt{d_k}} + M\right)V
\end{align}
Here, \( \sqrt{d_k} \) is the scaling factor, and \( M \) is a mask matrix enforcing causality, ensuring that tokens can only attend to previous or current tokens.
Specifically, \( M \) is defined as
\begin{align} 
M_{ij} &= 
\begin{cases}
        0, & \text{if } j \leq i \\
        -\infty, & \text{if } j > i.
    \end{cases}
\end{align}
This causal attention is illustrated in \cref{fig:2} (a).
\subsection{Three types of attention knockout.}
Attention knockout method \cite{geva2023dissecting} is a widely used technique in the field of large language model interpretability. It works by blocking the attention connections between specific tokens to study the impact of these particular information flows. After detailing the attention calculation in current Video-LLMs, we decompose the overall causal attention into three parts: Language-to-Video, Video Temporal attention, and Video Spatial attention. Correspondingly, we introduce three types of attention knockout to block specific attention flows, \ie, Language-to-Video Knockout (LV-K), Video Temporal Knockout (VT-K), and Video Spatial Knockout (VS-K), as shown in \cref{fig:2} (b), (c), and (d), respectively. Language-to-Video Knockout prohibits information flow from the video to the language. Video Temporal Knockout prevents information exchange between video frames, and Video Spatial Knockout prevents attention within each frame.
\begin{figure*}
    \centering
    \includegraphics[width=0.8\linewidth]{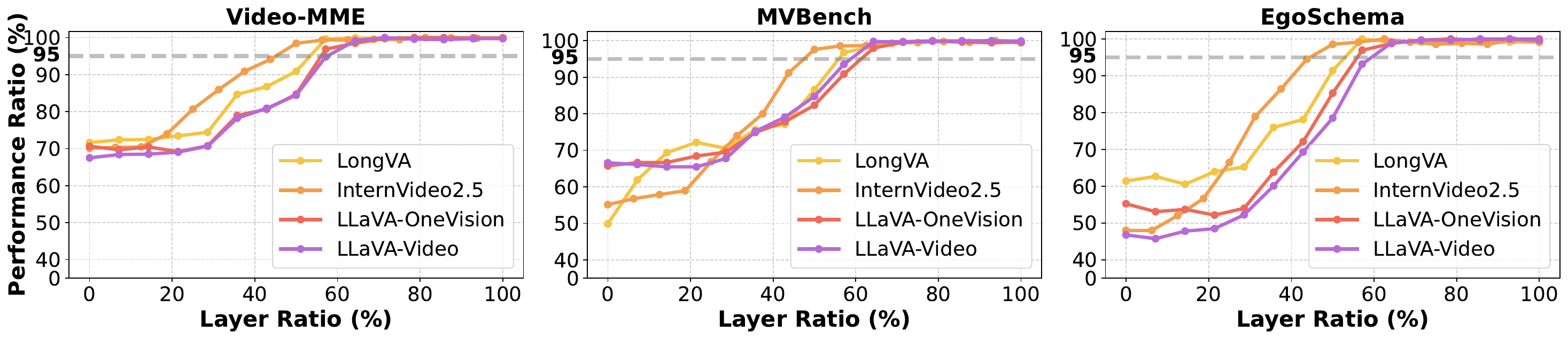}
    \caption{Normalized performance change on different benchmarks with different models. Setting: Apply Language-to-Video Knockout (LV-K) beyond a certain cutoff depth. For example, 60\% means that LV-K is applied to all layers beyond the first 60\% of layers. The normalization is with respect to the full model performance.}
    \label{fig:3}
    \label{fig:main_table_attention_knockout}
\end{figure*}

\begin{figure*}
    \centering
    \includegraphics[width=1\linewidth]{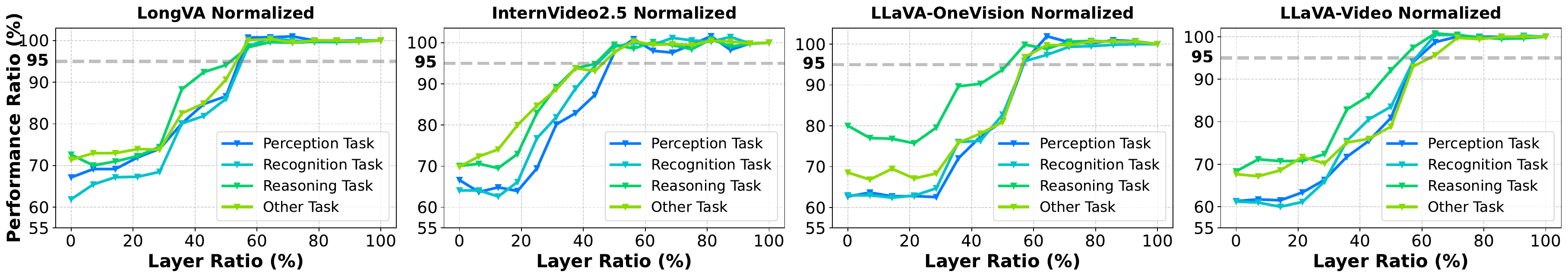}
    \caption{Normalized performance of different tasks on Video-MME using different models. Setting: Apply Language-to-Video Knockout (LV-K) beyond a certain cutoff depth. The normalization is with respect to the full model performance.}
    \label{fig:enter-label}
    \label{fig:4}
\end{figure*}

\subsection{Investigation Settings}

We adopt different settings for our investigation.  
Specifically, we consider two degrees of freedom: the layers where attention knockout is applied, and the corresponding knockout type (KT). For an $L$-layer Transformer, each layer's knockout configuration $L_i^{\text{KT}}$ has four possible options:
\begin{equation}
L_i^{\text{KT}} \in \{\text{no knockout},\, \text{LV-K},\, \text{VT-K},\, \text{VS-K}\}
\end{equation}
where $i \in \{1, \dots, L\}$. Here, no knockout represents the original attention operation. By carefully controlling these two variables, we define the following three investigation settings to explore answers to three questions about the internal mechanisms of Video-LLMs.

\noindent\textbf{Global Setting 1.}  
We block the language-to-video attention beyond a certain cutoff depth $i \in \{1, 3, 5, \dots, L\}$. The attention configuration of each layer $j$ is defined as:
\begin{equation}
L_j^{\text{KT}} =
\begin{cases}
\text{no knockout}, & j \leq i \\
\text{LV-K}, & j > i
\end{cases}
\label{eq:global1}
\end{equation}
This means the model can only access video information through the first $i$ layers. When $i = L$, no knockout is applied, serving as a baseline. We vary $i$ with a step size of 2 to track performance as a function of blocking depth. This setting is used to explore \textit{``Do Video-LLMs exhibit a clear staged pattern?''} as shown in \cref{fig:3,fig:4}.

\noindent\textbf{Global Setting 2.}  
To evaluate the overall contribution of different attention types, we select one type and apply the corresponding knockout to all layers. We iterate over all knockout types one by one:
\begin{equation}
L_j^{\text{KT}} = \text{KT}, \quad \forall j \in \{1, \dots, L\}
\label{eq:global2}
\end{equation}
where $\text{KT} \in \{\text{LV-K},\, \text{VT-K},\, \text{VS-K}\}$. This setting is used to explore \textit{``Globally, how does each attention type contribute to video QA performance?''} as shown in \cref{fig:5}.

\noindent\textbf{Fine-grained Setting.}  
To examine the impact of each attention type at a fine-grained level, we apply a specific knockout $\text{KT} \in \{\text{LV-K},\, \text{VT-K},\, \text{VS-K}\}$ within a sliding window of 4 layers ending at layer $x$ ($x \geq 4$), leaving layers outside this window without knockout. We define the affected layers as:
\begin{equation}
L_p^{\text{KT}} = \text{KT}, \quad \forall p \in \{x - 3,\, x - 2,\, x - 1,\, x\}
\label{eq:fine grained}
\end{equation}

This setting is used to explore \textit{``At a fine-grained level, how does each attention type impact video QA across different layers?''} as shown in \cref{fig:6}.

\section{Experiments and Observations}
In this section, we begin by outlining the datasets and models, followed by a presentation of the experimental results along with detailed analyses. % to demonstrate our findings. 
\subsection{Experiment details}
\begin{figure*}[t]
    \centering
    \includegraphics[width=1\linewidth]{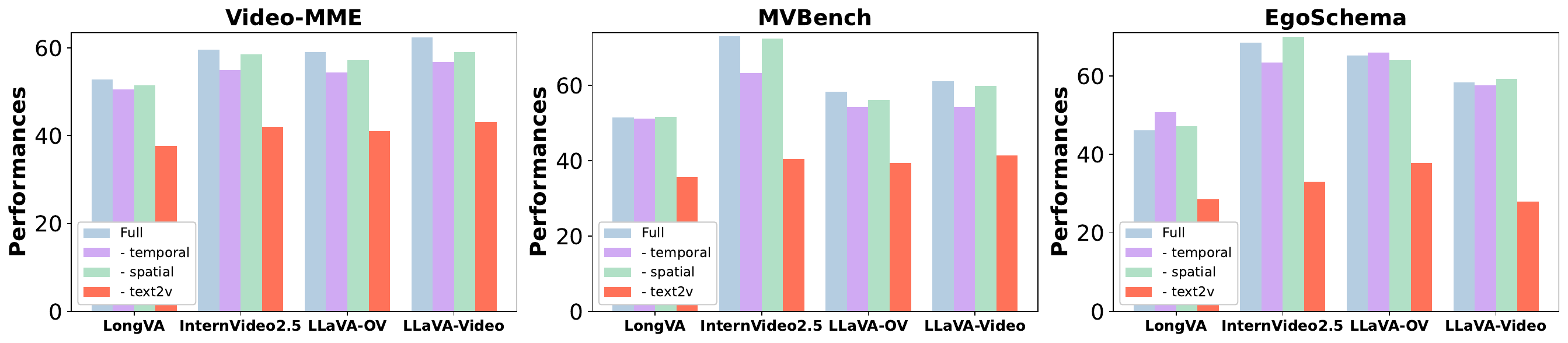}
    \caption{Absolute Performance on the different benchmarks. Setting: Applying different knockouts for all layers.
    }
    \label{fig:5}
\end{figure*}
\textbf{Datasets.} We focus on the video question answering task and use three well-established mainstream datasets (\ie, Video-MME \cite{fu2024video}, MVBench \cite{li2024mvbench}, and EgoSchema \cite{mangalam2023egoschema} to explore the internal mechanisms of Video-LLMs. These datasets differ in duration (ranging from 3 minutes to 1 hour), perspective (first-person and third-person views), scene context (\eg, household settings, movies), and question types (\eg, temporal and spatial reasoning). Video-MME \cite{fu2024video} consists of 900 videos totaling 254 hours, with 2,700 human-annotated QA pairs. The videos span six major domains—knowledge, film \& TV, sports, performing arts, daily life, and multilingual content—with varying video lengths. MVBench \cite{li2024mvbench} is a short-length multimodal benchmark designed to evaluate MLLMs' temporal reasoning in dynamic tasks. Unlike static image benchmarks, it includes 20 time-dependent tasks such as action sequences, prediction, object permanence, and motion analysis, each with 200 samples, totaling 4,000 QA pairs. EgoSchema dataset \cite{mangalam2023egoschema} comprises 5,000 multi-choice questions derived from 5,000 three-minute egocentric videos. It includes a publicly available subset of 500 questions, %with labels,
while the complete evaluation is on the server. Due to the large number of questions, we adopt the widely used subset for evaluation. \par\noindent\textbf{Models.} We investigate widely used open-source video large language models (Video-LLMs) \cite{zhang2024long, wang2025internvideo2, zhang2024llavanextvideo, li2024llava} that achieve cutting-edge performance across diverse video understanding tasks. These models employ similar architectures but are trained on different publicly available datasets, allowing us to systematically explore spatial-temporal modeling in Video-LLMs while minimizing confounding factors related to model architecture. We test four 7B models, including LongVA \cite{zhang2024long}, InternVideo2.5 \cite{wang2025internvideo2}, LLaVA-Video-7B \cite{zhang2024llavanextvideo}, and LLaVA-OneVision-7B \cite{li2024llava}. We also provide results from a larger 34B model in the Experiments section of the Appendix, while most experiments are conducted on 7B models due to computational cost constraints. For all models we test, we use the same uniform sampling strategy to sample 32 frames from each video. Other details are provided in the implementation details section of the Appendix.
\subsection{Do Video-LLMs exhibit a clear staged pattern?}\label{sec:early exit}
To investigate whether video processing in Video-LLMs also follows a clear staged pattern, we apply the Language-to-Video attention knockout to all layers after a certain layer, corresponding to Global Setting 1. Different Video-LLMs may vary in their performance across benchmarks and differ in the number of layers. For easier visualization and comparison, we introduce two metrics—\textbf{\textit{layer ratio}} and \textbf{\textit{performance ratio}}—to evaluate the impact of removing visual tokens beyond specific layers on model performance.
The \textbf{\textit{layer ratio}} represents the proportion of layers that retain language-to-video attention relative to the total number of layers. For example, a 60\% layer ratio means that the remaining 40\% of layers have Language-to-Video attention knocked out, preventing any further visual information from flowing to the text tokens.
The \textbf{\textit{performance ratio}} quantifies the relative performance of the model under the current knockout setting compared to its original (no-knockout) $100\%$ performance using $100\%$ layers. Fig.~\ref{fig:main_table_attention_knockout} shows the Video-LLMs’ performance across benchmarks.

\begin{figure*}[h!]
    \centering
    \includegraphics[width=1\linewidth]{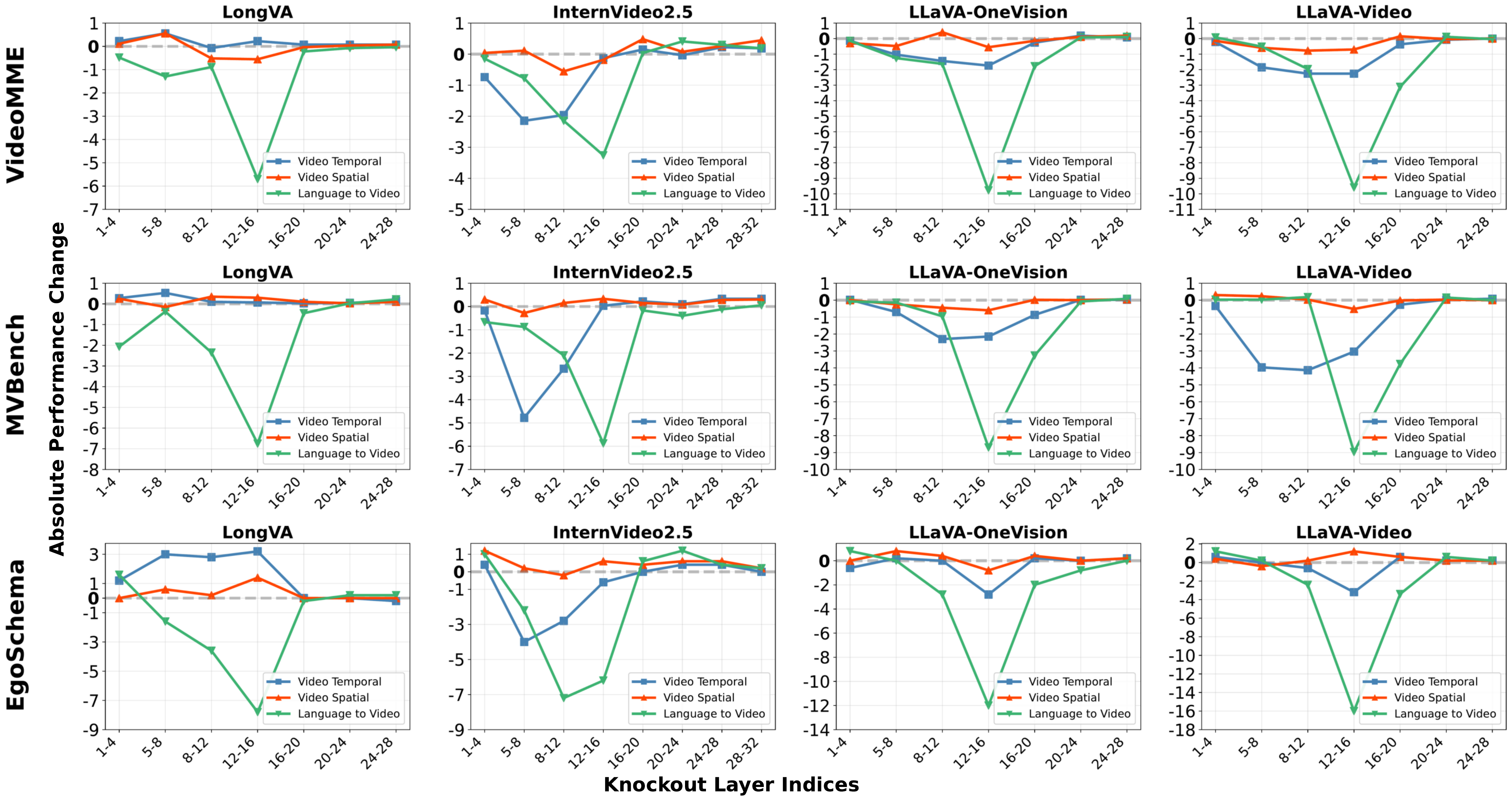}
        \caption{Absolute performance change across different benchmarks. Setting: applying different knockouts to layers within a sliding window.}
    \label{fig:6}
\end{figure*}

We draw the following observations:
(i) When all layers have no access to visual tokens, the three Video-LLMs retain at most 70\% of their original performance (e.g., LongVA on Video-MME) and as little as 48\% (e.g., LLaVA-Video on EgoSchema). This indicates that visual information plays a critical role in accurately answering questions, while also demonstrating the extensive world knowledge embedded in the language model.
(ii) As the layer ratio increases from 0\% to 60\%, model performance gradually improves, with the most significant boost occurring in the 50\%–60\% range across all video understanding benchmarks. This suggests that the middle layers contribute most to visual information processing, which is further supported by our fine grained setting experiments.
(iii) Blocking information beyond 60\% of the layer depth has little to no effect on model performance across different video understanding benchmarks. This suggests that visual information is primarily processed in the earlier layers, while the remaining layers are mainly responsible for higher-level reasoning—indicating a clear staged processing pattern. Moreover, Fig.~\ref{fig:4} illustrates the performance of different Video-LLMs across various tasks in the Video-MME benchmark, including perception, recognition, reasoning, and other task types. We observe that across different task categories, model performance exhibits a trend similar to that shown in Fig.~\ref{fig:main_table_attention_knockout}, as visual tokens are progressively blocked. This further validates the effectiveness of our findings. Additional task-level performance results of different Video-LLMs on various benchmarks, qualitative results and a a larger model with 32 billion parameters experiments also observe a similar pattern, which further confirms the existence of this trend. which can all be found in the experiments section of the Appendix.

\begin{table*}[htbp]
\centering
\begin{tabularx}{\textwidth}{c>{\centering\arraybackslash}X>{\centering\arraybackslash}X>{\centering\arraybackslash}X>{\centering\arraybackslash}X>{\centering\arraybackslash}X>{\centering\arraybackslash}X}
\toprule
Model & Settings & Attention Flops& MME & MVBench & EgoSchema \\
\midrule
\multirow{2}{*}{Llava-Video} %169
& Baseline  &100\%   & 62.4 & 61.1 & 58.4  \\
& \textit{Exit only} &64.3\% & 62.0 & 61.1 & 58.0  \\
& \textit{Exit + window}& 37.1\%  & 60.0 & 60.8 & 58.2  \\
\midrule
\multirow{2}{*}{Llava-OneVision} %196
& Baseline   &100\%  & 59.1 & 58.3 & 65.2  \\
& \textit{Exit only}  &64.3\% & 58.2 & 57.3 & 64.4  \\
& \textit{Exit + window}& 37.1\%  & 58.0 & 57.6 & 65.2  \\
\bottomrule
\end{tabularx}
\caption{Performance of different models under various settings on three benchmarks. \textit{Exit only} means that video tokens exit the model after a certain layer. \textit{Exit + window} means that, in addition to exiting, we also control the temporal attention range—i.e., for certain layers, video frames are only allowed to perform spatial attention. For both LLaVA-OneVision and LLaVA-Video, we exit video tokens after layer 18 and limit the first 8 layers to perform spatial attention only, as they act as non-critical layers.}
\label{tab:1}
\end{table*}

\subsection{Globally, how does each attention type contribute to video QA performance?} \label{sec:how spatial-temporal}
We further explore how each type of attention contributes to video question answering performance from a global view. To answer this, we conduct experiments under Global Setting 2. Specifically, we systematically apply each type of attention knockout to all layers of each model and evaluate their performance on various benchmarks. As shown in Fig.~\ref{fig:5}, applying video temporal knockout and video spatial knockout across all layers results in minimal performance degradation across different benchmarks. However, applying language-to-video knockout across all layers leads to a significant drop in performance. This indicates that spatial-temporal modeling in Video-LLMs primarily occurs through the interaction between language tokens and video tokens, while temporal and spatial self-attention contribute substantially less. Notably, in VideoQA tasks, the computational cost of temporal and spatial attention is often much higher than that of language-to-video attention while the current video relies much more on the language-to-video attention instead of other two.
\subsection{At a fine-grained level, how does each attention type impact video QA across different layers?} We conduct experiments under fine grained Setting, we adopt each type of knockout to a small set of layers in a sliding window (we use 4), then examine inside this speicifc window how each type attention contribute to the final answer. Here we report the absolute performance change of these tested models on each task dataset.
As shown in Fig.~\ref{fig:6}, we observe the following:
(i) For most sliding windows of layers, language-to-video attention knockouts lead to a significantly larger performance drop compared to temporal or spatial attention knockouts, as shown in \cref{fig:6}.
(ii) For a subset of layers, applying knockouts causes a substantial performance drop, while knockouts on the remaining layers have only minor effects.
(iii) For most individual layers, language-to-video attention knockouts have a stronger impact than either temporal or spatial attention knockouts.
(iv) In some cases, applying knockouts even improves performance. For instance, LongVA shows improved performance on EgoSchema when certain layers are knocked out.

\section{Potential Applications}
Our previous observations reveal the inefficiency of existing Video-LLMs on VideoQA tasks. For example, in the global setting, we identified a two-stage processing pattern in Video-LLMs. Specifically, blocking all visual tokens in the second stage can nearly preserve the original performance while significantly reducing computational costs. Taking LLaVA-Video as an illustrative example, each video frame contains 196 tokens, and with 32 frames per video, the full video sequence length reaches \(196 \times 32\) tokens. In contrast, text sequences typically contain fewer tokens than a single frame (196 tokens). Consequently, video attention computation per layer is more than \(32^2\) times higher than that of language self-attention. Thus, substantial computational savings can be achieved by omitting video tokens in the second stage. In our fine-grained setting, we discovered distinct layer-level outliers—a property that can be leveraged to further reduce first-stage computation. Specifically, temporal attention incurs a computational cost 31 times higher than spatial attention. By constraining attention within each frame to spatial attention prior to these identified outliers, we substantially reduce the overall computational load. We propose a straightforward strategy: for LLaVA-OneVision and LLaVA-Video, we block temporal attention from layers 1 to 8 (with layer 8 identified as an outlier), followed by the removal of visual tokens from layer 18 onwards. Results presented in \cref{tab:1} demonstrate that this strategy maintains performance comparable to baseline methods while substantially reducing computational overhead. The detailed FLOPs calculations and results for LongVA and InternVideo2.5 are provided in the experimental section of the Supplementary Material.

\section{Conclusion}
In this paper, we reveal the internal working mechanisms of Video-LLMs when handling VideoQA tasks. Our experiments demonstrate that different Video-LLMs exhibit similar processing patterns across various benchmarks and tasks. Specifically, video information extraction is completed in the early layers, while video reasoning occurs in the later layers. Moreover, spatial-temporal modeling is primarily driven by language-guided retrieval, rather than by Video Temporal or Video Spatial attention mechanisms. Also, a small subset of layers plays a critical role in VideoQA. Finally, we show that computational cost can be reduced by minimizing the computation in non-critical layers and by exiting video tokens in the second stage. These findings enhance the interpretability of Video-LLMs, providing new research directions for a deeper understanding of how these models process and comprehend video content. Additionally, they offer insights into improving effectiveness and efficiency in downstream tasks and optimizing model design.

\small
\bibliography{video2025}

\end{document}